# The Potential of Convolutional Neural Networks for Cancer Detection

Hossein Molaeian, Kaveh Karamjani, Sina Teimouri, Saeed Roshani, Sobhan Roshani
Department of Computer Engineering, Islamic Azad University – Kermanshah Branch, Kermanshah, Iran

**ABSTRACT:**

Early detection is crucial for successful cancer treatment and increasing survivability rates, particularly in the most common forms. Ten different cancers have been identified in most of these advances that effectively use CNNs (Convolutional Neural Networks) for classification. The distinct architectures of CNNs used in each study concentrate on pattern recognition for different types of cancer across various datasets. The advantages and disadvantages of each approach are identified by comparing these architectures. This study explores the potential of integrating CNNs into clinical practice to complement traditional diagnostic methods. It also identifies the top-performing CNN architectures, highlighting their role in enhancing diagnostic capabilities in healthcare.

## 1. INTRODUCTION

### 1.1. Cancer

Cancer is one of the most complex and deadly diseases of the present century, and its increasing prevalence has turned it into a global crisis. The disease is characterized by the uncontrolled proliferation of cells that can spread throughout the body, often leading to disability and death. The causes of cancer vary and arise from a combination of genetic, environmental, and lifestyle factors.[1]

This study emphasizes the most common types of cancer, including prostate cancer, blood cancers (such as leukemia and lymphoma), bladder cancer, skin cancer (both melanoma and non-melanoma), colorectal cancer, liver cancer, breast cancer, ovarian cancer, thyroid cancer, and lung cancer.

Early cancer detection significantly increases the chances of recovery and reduces associated mortality. Early-stage cancers often may not present obvious symptoms and can be detected incidentally during routine check-ups. Early detection allows for treatments that are more likely to be successful and have fewer side effects. Imaging plays a crucial role in cancer diagnosis, helping doctors determine the tumor's size, shape, and location, as well as how far it may have spread.

The imaging techniques primarily include radiography using X-ray to capture images of bones and nearby soft tissues; ultrasound, which employs high-frequency sound waves to create pictures of low-density soft organs; CT scans, which merge X-ray technology with computer technology for cross-sectional views of the entire body; MRI, utilizing mid-radio and high-frequency energy to achieve high sensitivity and resolution of soft tissues; and finally, PET scans, which involve injecting radioactive substances into the patient to visualize cellular metabolic activities over time. In Figure 1, we have 4 cancer photography models as examples.

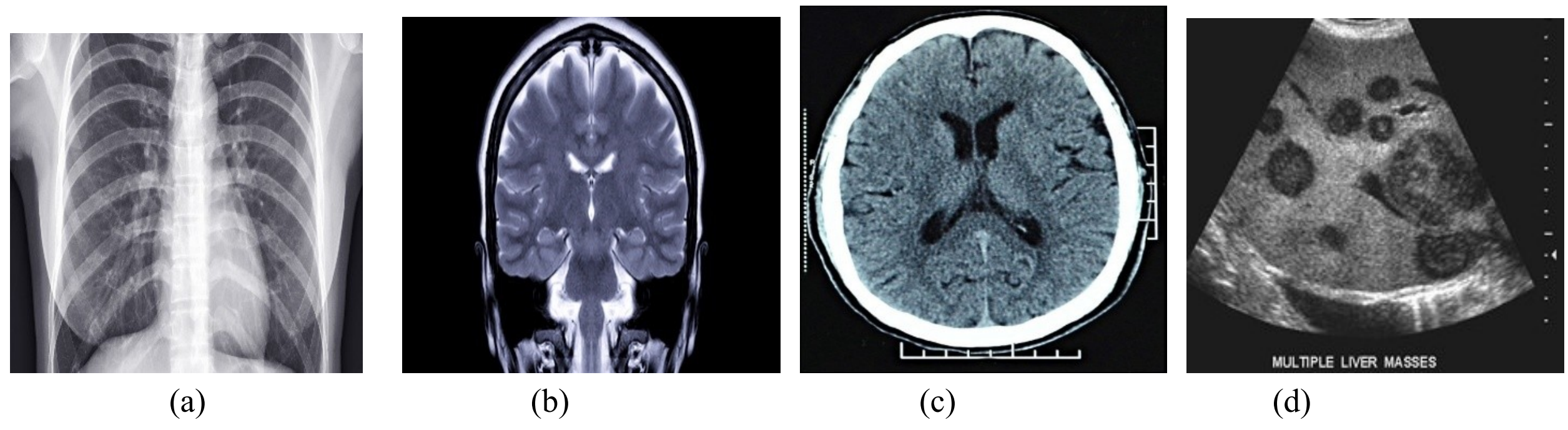

***Fig1.*** *Here are some examples of imaging techniques used for cancer diagnosis, with the labels assigned to each image: (a) Radiography, (b) Ultrasound, (c) CT scan, and (d) MRI*[2,3]

We have compiled comprehensive information in the form of tables and charts, allowing us to visualize the mortality rates and the percentage of cases for each type of cancer as demonstrated in Table 1 and Figure 2. These insights help identify which cancer has the highest mortality rate and enable potential for improvements in diagnostic methods to reduce these percentages in future research.

***Table 1.*** *This table provides an overview of various types of cancer, including a brief description of each type and the annual number of deaths and new cases reported.*[4-7]

| Cancer Type | Annual Incidence Rate | Annual Mortality Rate | Annual Growth Rate (%) | Description |
|---|---|---|---|---|
| **Prostate** | 1.4M | 375,000 | 2.0% | One of the most common cancers in men, which develops in the prostate gland and grows slowly |
| **Leukemia** | 476,000 | 311,000 | 1.5% | Cancer usually begins in the blood cells and bone marrow, increasing abnormal blood cells. |
| **Bladder** | 573,000 | 213,000 | 2.2% | Most often, it develops in the inner lining of the bladder and is more common in men. |
| **Skin** | 325,000 | 57,000 | 3.0% | A type of cancer that begins in the melanin-producing cells of the skin and can spread to other parts of the body. |
| **Colorectal** | 1.9M | 935,000 | 1.8% | One of the common cancer and The cells lining the colon or rectum become abnormal and grow out of control. |
| **Liver** | 900,000 | 830,000 | 2.5% | Often develops as a result of hepatitis or other |

| | | | | liver damage and progresses rapidly. |
|---|---|---|---|---|
| **Breast** | 2.3M | 685,000 | 2.6% | The most common cancer among women begins in the cells of the milk glands or ducts of the breast. |
| **Ovarian** | 313,000 | 207,000 | 1.4% | A type of cancer that begins in the ovaries and is often diagnosed at an advanced stage. |
| **Thyroid** | 587,000 | 43,000 | 2.0% | A cancer that occurs in the thyroid gland is more common in women and is often treatable with surgery. |
| **Lung** | 2.2M | 1.8M | 1.9% | One of the deadliest cancers, which develops in the lungs and is primarily caused by tobacco use. |

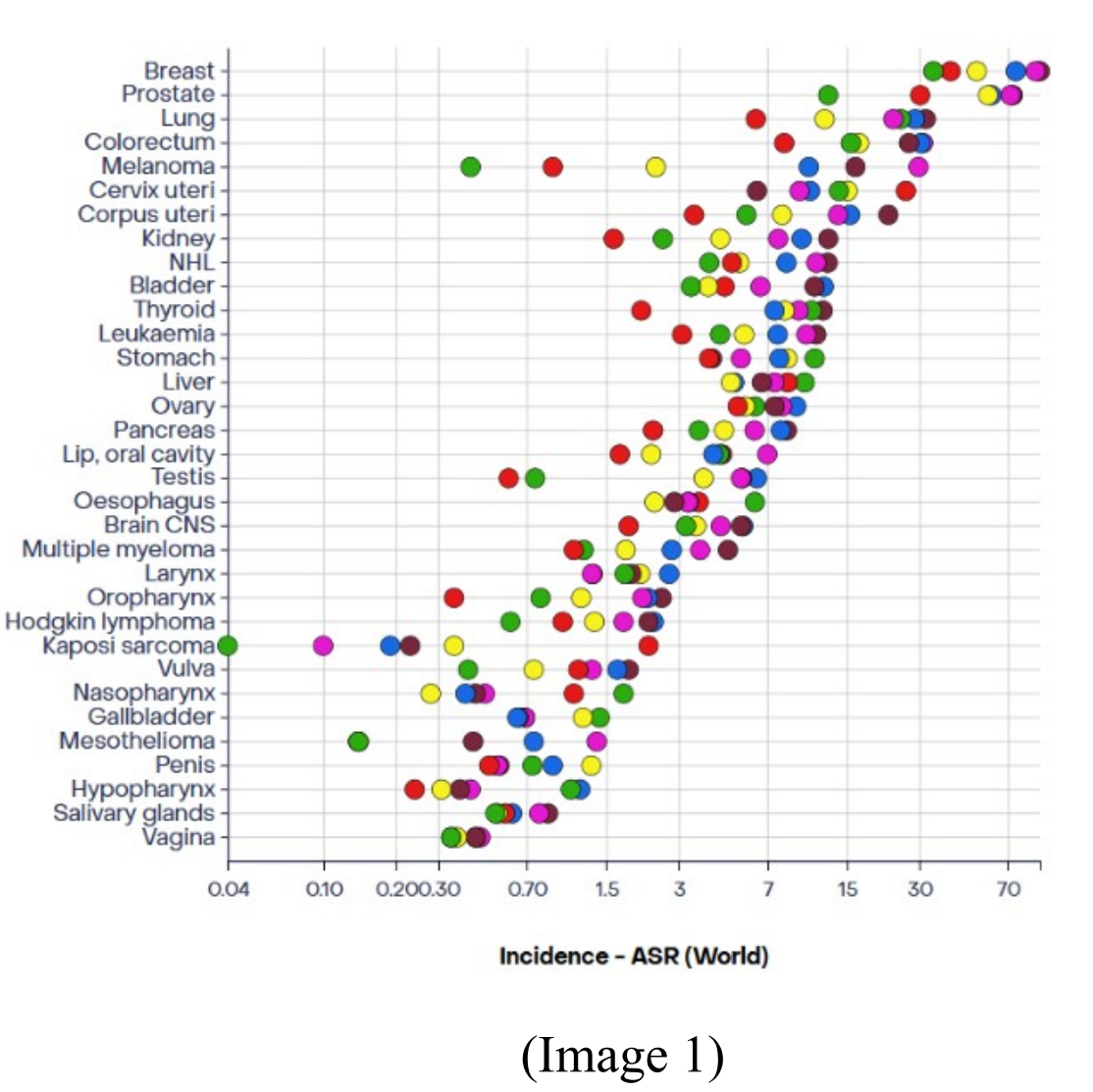

(Image 1)

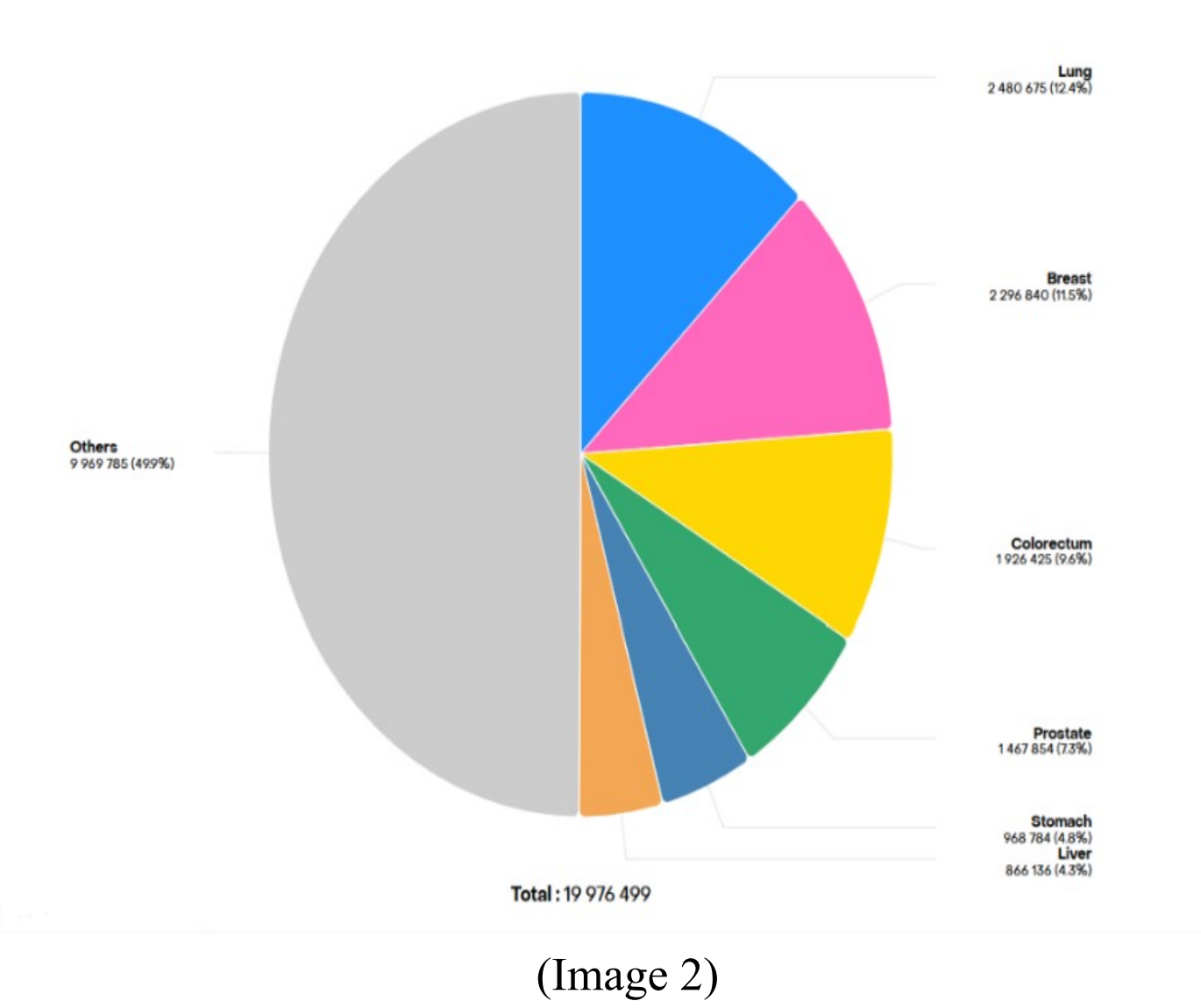

(Image 2)

***Fig 2.*** *illustrates the global age-standardized incidence rates (ASR) of various cancer types, highlighting disparities across different forms and regions. Image 2 displays the international distribution of cancer cases by type, with lung, breast, and colorectal cancers making up the most significant shares.*[7]

### 1.2. **CNNs:**

CNNs are vital in image processing and pattern recognition, making them one of the most important and commonly used models in artificial intelligence.[8] CNNs are highly effective in identifying and classifying images and have numerous applications in medical diagnosis.[9,10]

The CNN model is organized in several layers, such that higher-order features of an image are interpreted in a progressing hierarchy. The core of the architecture involves convolutional layers, pooling layers, and fully connected layers.[11] ***Fig3***

Convolutional Layer: This core layer learns features from input images using filters or kernels that slide across the image, detecting patterns like edges and textures.

Pooling Layer: This layer reduces the image size after convolution, decreasing computational load and improving efficiency. Max Pooling is a common technique, extracting the maximum value from small sections.

Fully Connected Layer: In this layer, features from previous layers are classified into meaningful categories.

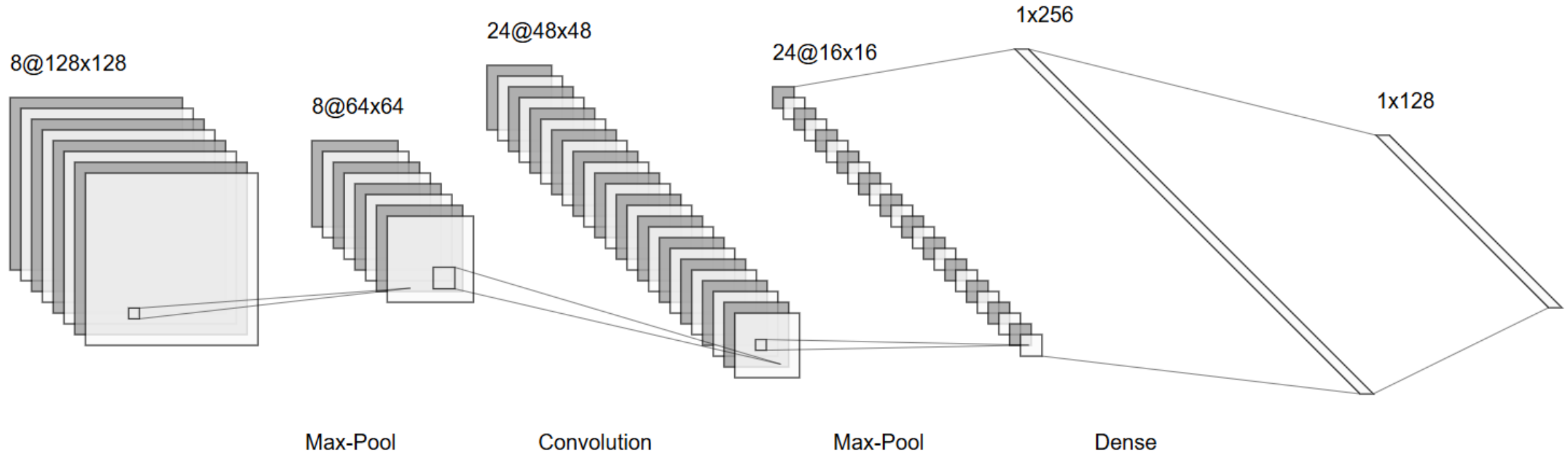

***Fig3.*** *CNNs' architecture consists of convolutional, max-pooling, and dense (fully connected) layers. Each layer sequentially extracts more complex features from the image data.*[12]

The entire learning process in a neural network rests on mathematical computations that permit analysis of input data to achieve an output.[13,14]

Differentiation and optimization are among the main principles behind neural networks. Differentiation provides a way to compute the minimum error function that leads the network to peak performance. The error or loss function measures how well a network predicts by allowing optimization algorithms to minimize that particular function.[15]

Essential features are extracted from the image using filters, and the application of convolution thereby establishes such extraction. When these formulas come together, the model can optimize the learning process and further bias toward accuracy over time.[16]

Weighted Sum and Bias:

$$z = w_1 x_1 + w_2 x_2 + \cdots + w_n x_n + b \qquad (1)$$

(x) the input values are denoted by.

(w) represents the weights corresponding to each input.

(b) is the bias (a constant value) that helps the model have more flexibility.

Convolution in CNNs:

$$S(i,j) = (I * K)(i,j) = \sum_m \sum_n I(i+m, j+n) \cdot K(m,n) \qquad (2)$$

(I) is the input image matrix.

(K) is the kernel or filter matrix.

(S(i, j)) is the output value at the point ((i, j)).

## Loss Function:

The loss function measures a model's predictive performance, quantifying the deviation from the actual outcomes.

$$Loss = -[y \cdot \log(\hat{y}) + (1-y) \cdot \log(1-\hat{y})] \quad (3)$$

($\hat{y}$) is the model's prediction (the predicted probability of cancer, a value between 0 and 1).

## Cost Function:

The mean or cumulation of the model's errors from all samples.

$$Cost = \frac{1}{N} \sum_{i=1}^{1} Loss_i \quad (4)$$

(N) is the number of samples (e.g., the number of images)

($Loss_i$) is the error for each sample.

**Kernels in CNNs:**

Kernels (or filters) are essential components of a convolutional neural network (CNN) that detect and extract patterns from images. In small matrices (3x3 or 5x5), kernels move across the image and perform convolution, involving element-wise multiplication followed by summation to extract features.[17]

In Figures 4 and 5, we can observe how the kernel operates which provides a better understanding of the functionality of filters in neural networks.

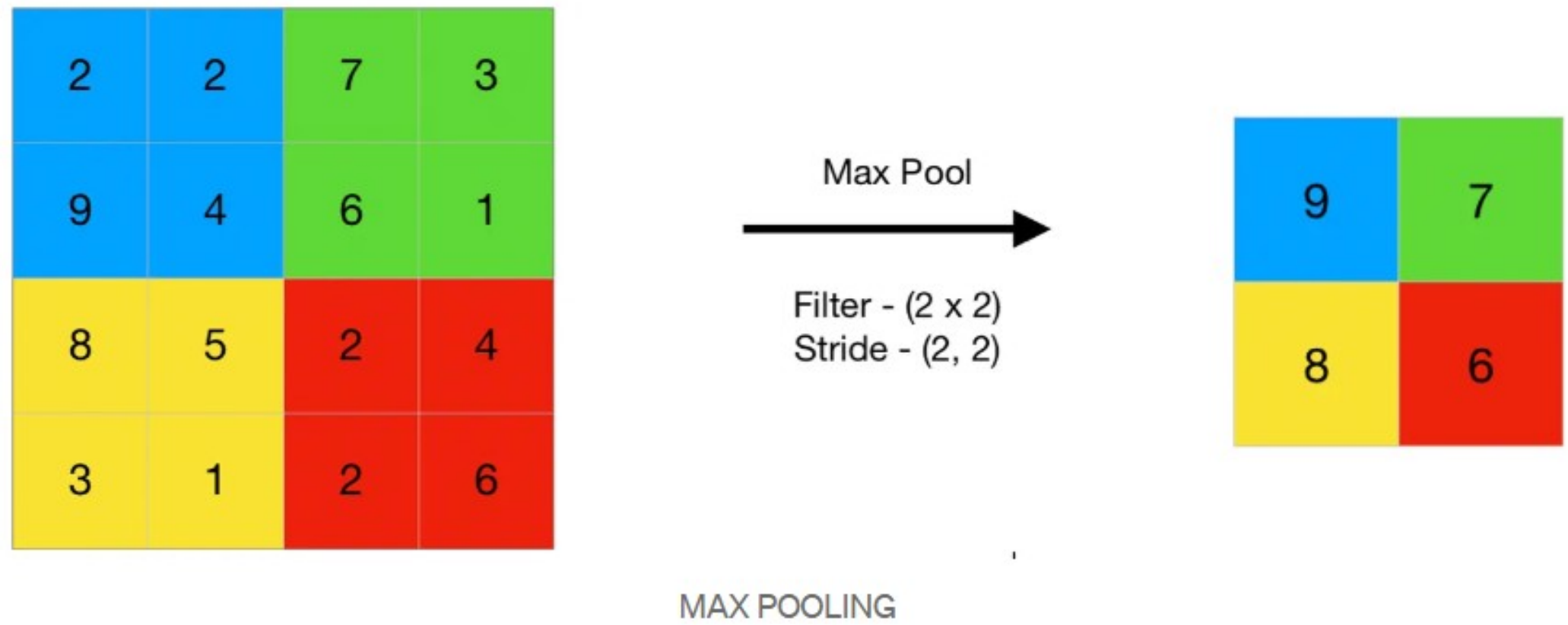

***Fig4.*** *This image illustrates the Max Pooling process in neural networks. A 4×4 input matrix is processed using a 2×2 filter with a stride of 2. For each 2×2 block, the maximum value is selected, resulting in a 2×2 output matrix.*[18]

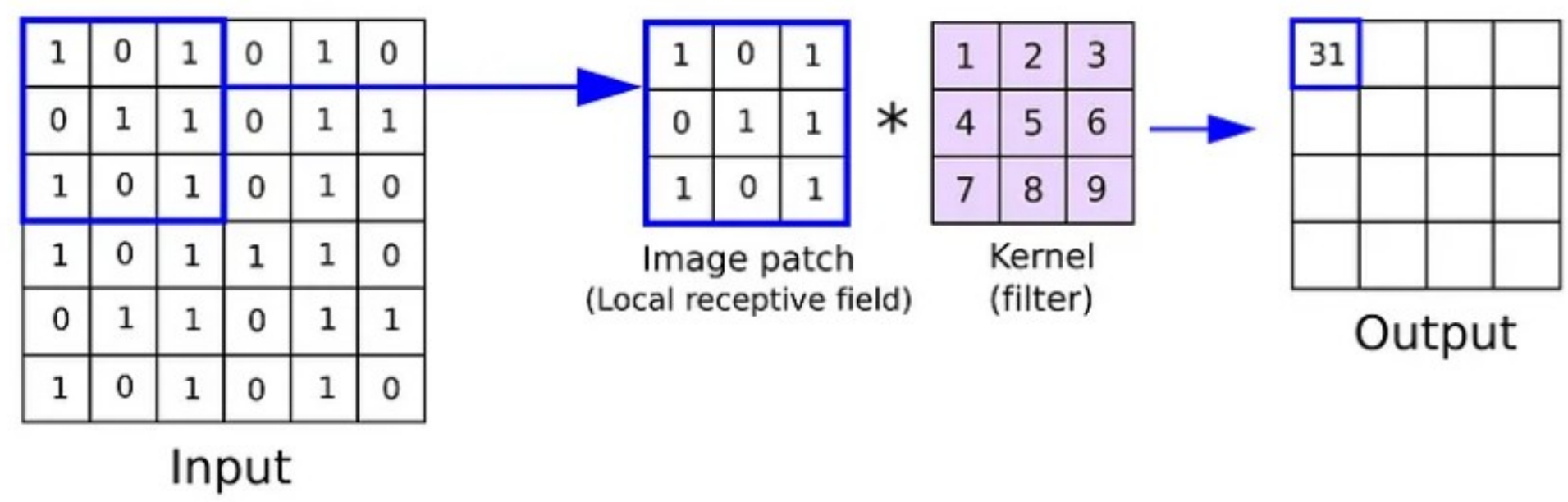

***Fig5.*** *This image illustrates the convolution process in neural networks. A small region of the input matrix is selected and multiplied element-wise with a kernel. The resulting products are summed up, and a single value is stored in the output matrix.*[18]

CNNs analyze medical images to identify patterns indicating the presence of cancer cells or damaged tissue. They are trained on large datasets that include images from samples affected by cancer and healthy organs, learning to recognize the features associated with those cancerous cells and using this information to diagnose new cases.[19]

Although CNNs show great promise in cancer detection, challenges persist regarding access to high-quality data that lacks sufficient homogeneity. Poor data can lead to incorrect conclusions. Additionally, interpreting the results may be equally challenging since the technical complexity of the CNN structure requires a deep understanding of both medicine and artificial intelligence.[20,21]

## 2. METHODOLOGY

The articles selected for this review were based on specific criteria to guarantee a comprehensive exploration of CNN applications in cancer detection. The emphasis was on their effectiveness in analyzing and interpreting medical images. Considering the significant advancements in artificial intelligence in recent years and the resulting improvements in the accuracy and speed of CNN-based diagnostics, only articles published after 2020 were considered. This strategy ensured that only the most current and reliable studies were incorporated, concentrating on recent methods in the field.

Selecting appropriate articles involved consulting reputable scientific databases, primarily PubMed, IEEE, and other research platforms known for their broad reach and reliable quality in artificial intelligence and medical research. Due to various access restrictions on many papers, the selection was limited to free or open-access documents. Although this made the article selection process more challenging, it resulted in a solid collection of credible research from trusted sources.[22]

Out of the approximately 20 articles initially selected and reviewed based on the above criteria, 10 top articles were ultimately chosen. These 10 articles not only focused on cancer detection but also covered various common and well-known types of cancer that hold significant clinical and medical importance. Given the lethal nature and diagnostic complexity of cancer, this selection allowed us to concentrate on research related to cancers analyzed with CNNs in a more detailed manner and explore each study's different aspects.

This study further examines how CNNs collaborate with various types of medical imaging to produce faster and more accurate prognostic predictions in cancer detection. The challenges and limitations associated with implementing CNNs in this area have been thoroughly discussed, particularly concerning the increased computational power required and the lack of diverse, high-quality datasets suitable for these applications.

## 3. Analysis and Comparison:

### 3.1 Prostate Cancer:

This work presents an MRI dataset from Harvard, consisting of 482 prostate cancer patients and 200 brachytherapy cases, which have been divided into 70% for training and 30% for testing validation/testing.[23] Using a CNN and Transfer Learning Model[24,25] The diagnostic process was simplified. Additionally, various methods, such as Decision Trees and Support Vector Machines (SVM), were employed, with the best result achieved using the GoogLeNet model.[26] Although other methods were less successful, the SVM model achieved an accuracy of 99.71%. The losses were 0.7086 with mini-batch processing and 0.4206 during validation for the model's initial run, which was steadily reduced to nearly zero after 30 iterations. By the end of its 45th iteration, the model reached a validation accuracy of 100% based on its dataset.[27,28]

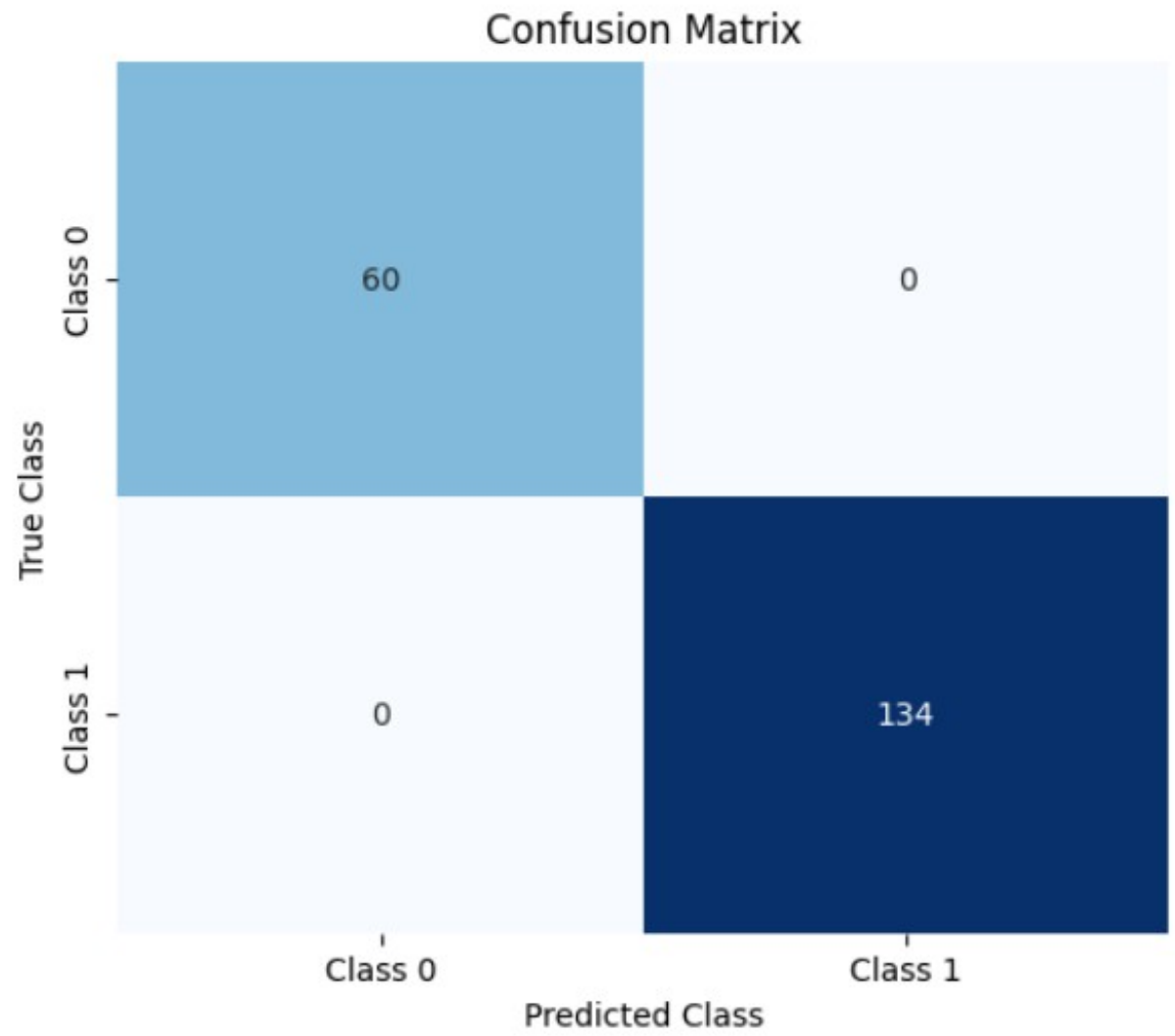

***Fig6.*** *Confusion matrix showcasing the performance of the CNN model in detecting prostate cancer, emphasizing its classification accuracy and false-positive rates.*

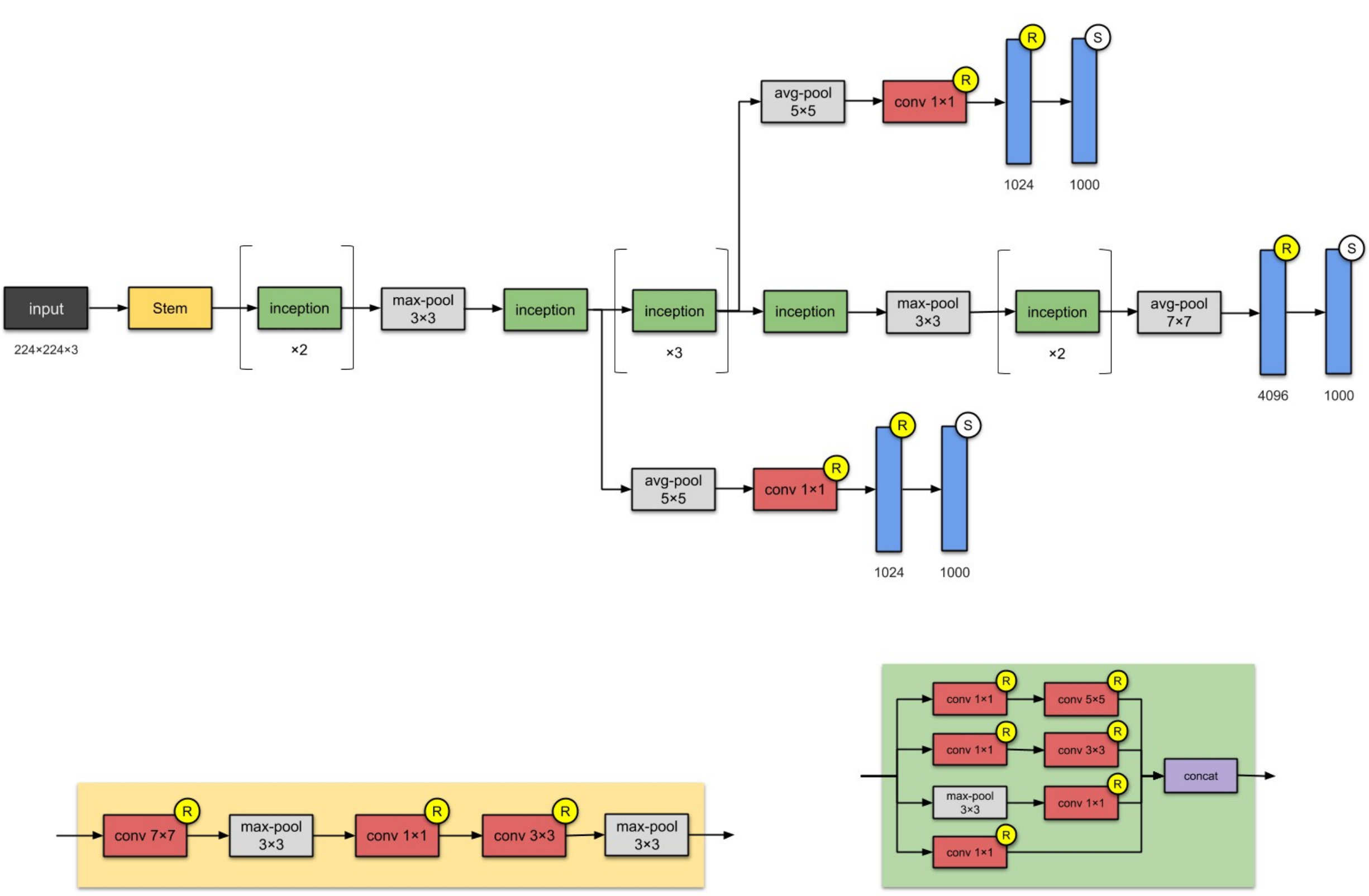

***Fig7.*** *Architectural illustration of the GoogLeNet model used in prostate cancer detection, detailing layers and connections.*[29]

## 3.2 Lung Cancer:

The proposed research will introduce the Iraqi Teaching Hospital Cancer Dataset, IQ-OTH/NCCD, which includes patients with various types of lung cancer and several individuals with healthy lungs, as shown in the CT scans.[30,31] The dataset comprises 1,190 images categorized as benign, malignant, and routine. The septum does not cover that of the last study; instead, it incorporates the AlexNet model. The training-testing ratio is 70-30. The model trained itself randomly throughout the entire training phase without any variation, achieving an overall accuracy of 93.548% at 86 of the total 100 epochs. The precision reached was 97.1015%, sensitivity was 95.714%, and specificity was 95%.[32]

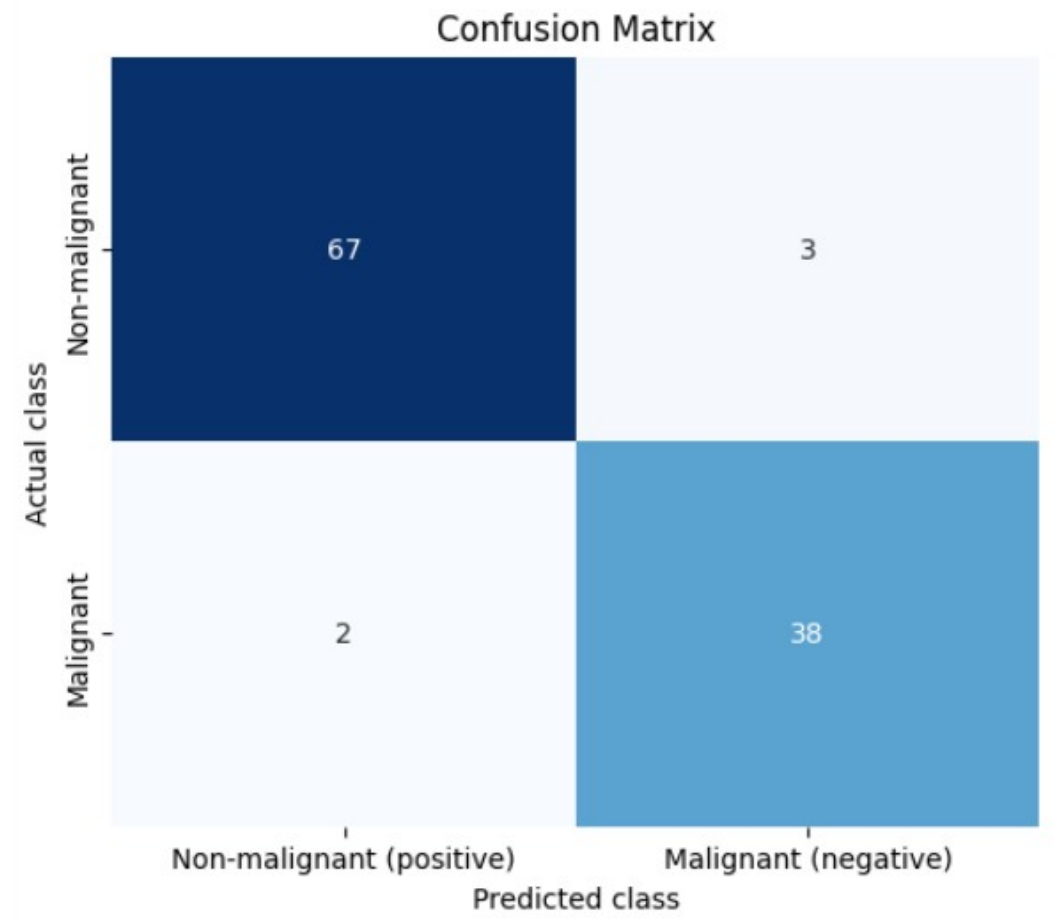

***Fig8.*** *Confusion matrix for the CNN model applied to lung cancer detection, highlighting precision, sensitivity, and specificity metrics.*

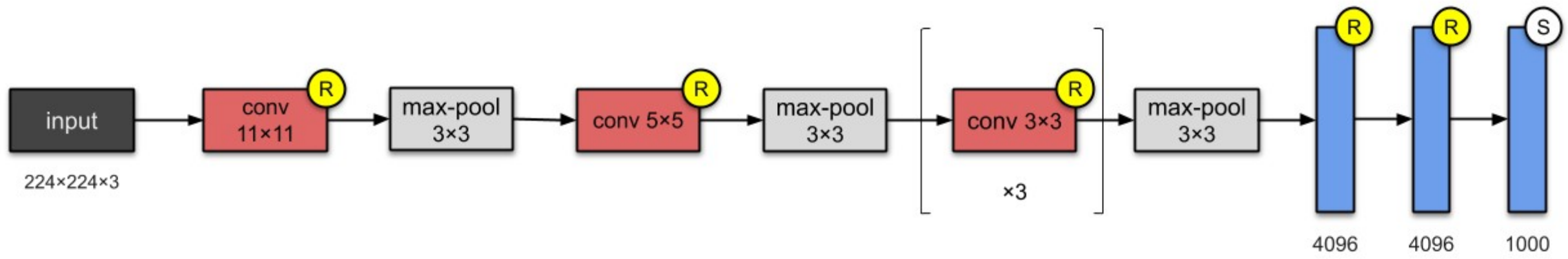

***Fig9.*** *Schematic representation of the AlexNet architecture, optimized for lung cancer classification, showing layers and kernel sizes.*[33]

### 3.3 Leukemia:

The authors moved away from the pre-trained models or Transfer Learning approaches used in the first two papers, creating an adapted version of the BCNN framework. The study included 260 microscopic images showing lymphocytes, some cancerous and others non-cancerous. The images are categorized into two classes: one consists of cancerous lymphocytes, while the other contains non-cancerous cells. Each class comprises 100 training images, 15 validation images, and 15 test images. This is accomplished using the Hold-out method process.[34]

Different models were examined using various combinations of convolutional and hidden layers. The model with three convolutional layers and five hidden layers performed exceptionally well, achieving an average test accuracy of 94.00% and a standard deviation of 2.00%. It had a 20% dropout rate on the convolutional layers, while a 50% dropout rate was applied to the fully connected layers to address potential overfitting on the evenly distributed data. A learning rate of 0.0001 using RMSprop was utilized to optimize the model over 50 epochs during the training phase, where dropout became one of the strategies ultimately employed to mitigate overfitting.

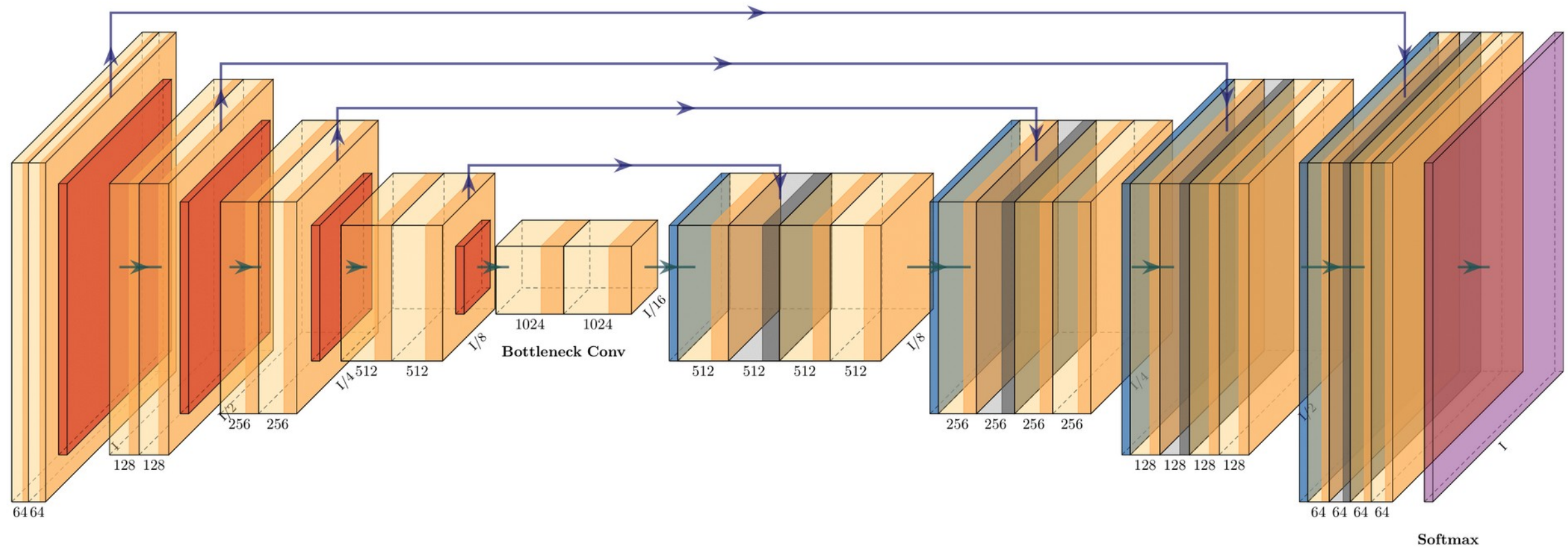

***Fig10.*** *Detailed architecture of the custom BCNN model used for leukemia diagnosis, illustrating its convolutional and fully connected layers*

### 3.4 Bladder Cancer:

More than 1.2 million publicly available images were utilized to train the model on the ImageNet dataset for pretraining. This dataset included images of 1,671 natural textures and 431 tumor lesions. The preprocessing was minimal, and the photos were stored as TIFF files.[35] It was further trained with 8,728 gastroscopy images to identify relevant texture features. Finally, transfer learning was applied to optimize the model using 2,102 cystoscopy images, including standard and tumor samples.[36]

Fifty-three observers with varying experiences assessed the AI model's performance against pathological findings and the Youden index. In this stepwise organ transfer learning approach, the CNN model evaluation demonstrated a sensitivity of 95.4%, specificity of 97.6%, AUC of 0.98, and Youden's index of 0.930. The diagnostic accuracy of observers, including medical students and urologists, was exceeded when tumors occupied more than 10% of the image. The model diagnoses in 5 seconds and takes 10 minutes to train.[37]

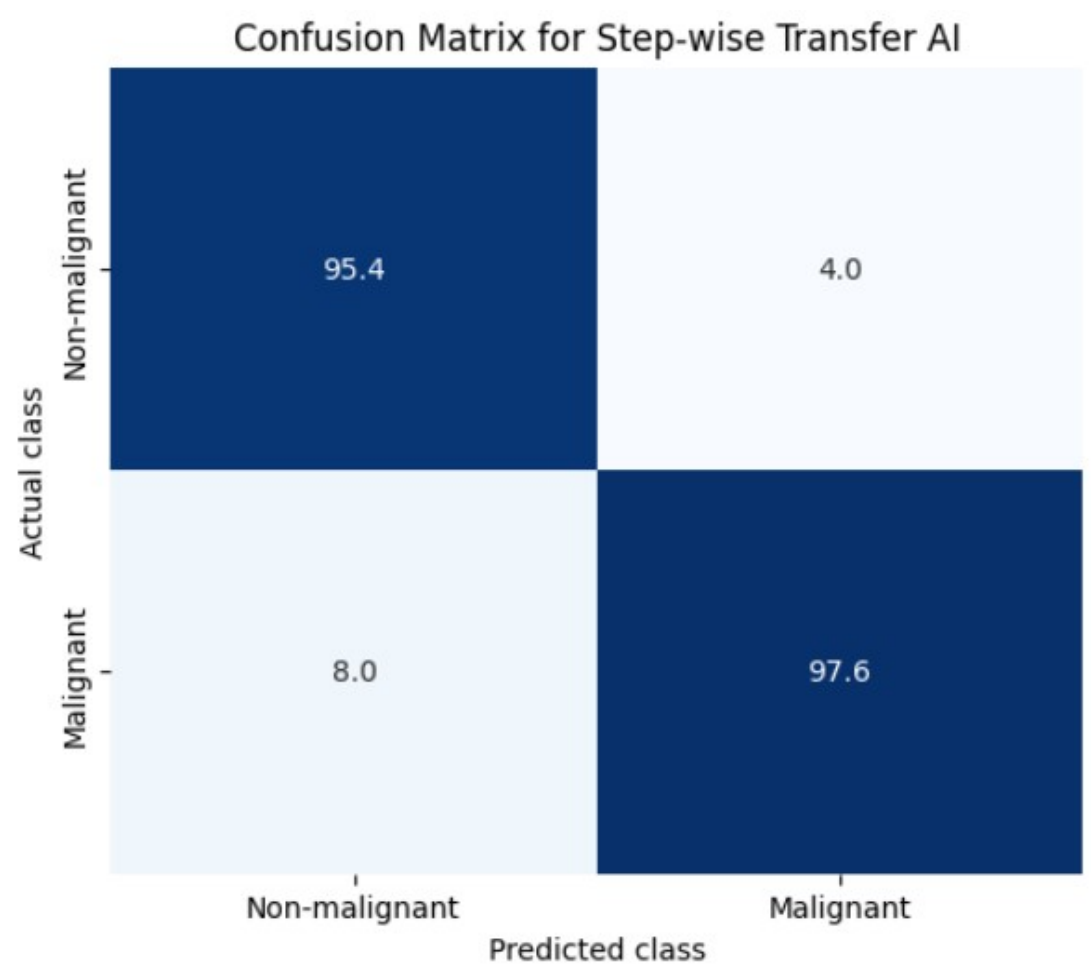

***Fig11.*** *Confusion matrix for bladder cancer detection, showcasing the CNN's diagnostic sensitivity and specificity metrics.*

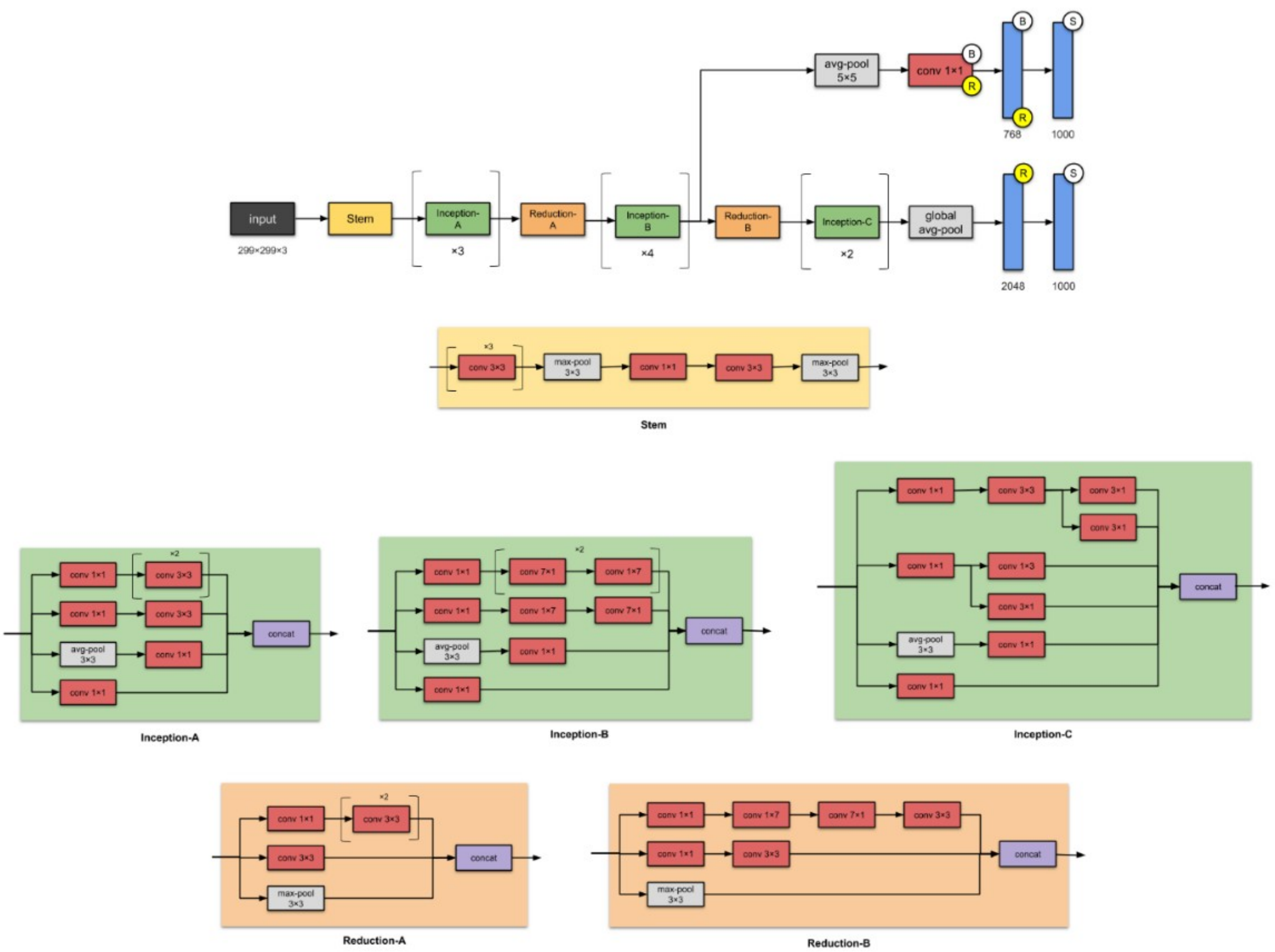

***Fig12.*** *Diagram of the CNN architecture applied to bladder cancer classification, featuring its layer structure and pooling mechanisms.*

## 3.5 Melanoma:

A dataset of approximately 800 images of four types of skin cancers—actinic keratosis, basal cell carcinoma, malignant melanoma, and squamous cell carcinoma—was created. Various augmentation strategies increased the total number of images to 5,600. Finally, a deep convolutional neural network was developed to train on this dataset for the specified images.[38] This proposed model is optimized using a learning rate of 0.001 and the Cross-Entropy loss function with an Adam optimizer. Subsequently, the model achieved an impressive 95.98% on the test data, surpassing two pre-trained models: GoogLeNet and MobileNet. The model exceeded GoogLeNet by 1.76% and surpassed MobileNet by approximately 1.12% while maintaining a computational complexity comparable to other models.[39] These studies evaluated InceptionV4 using classifiers such as SVM and Random Forest but achieved an unsatisfactory accuracy of approximately 89%. Modified ResNet-50 also reported its classification accuracy on skin lesions, reaching 85.8% for single images and 86.6% for multimodal networks; however, efforts to obtain higher accuracy were unsuccessful.[40] These 48-layer and 28-layer models were pre-trained on ImageNet with 1,000 classes, producing satisfactory results. In our experiments, we used these models along with their pre-trained weights. The dimensions of our images were 299×299 pixels. We also utilized the SGD optimizer and the Binary Cross-Entropy loss function.[41] Among all the classes, the highest accuracy in model performance was 97.23% for Basal Cell Carcinoma (BCC), with corresponding precision, recall, and NPV metrics optimizing for performance. The model achieved an accuracy of 95.98%, with precision and recall values of 91.96% and 91.97%, respectively.[42]

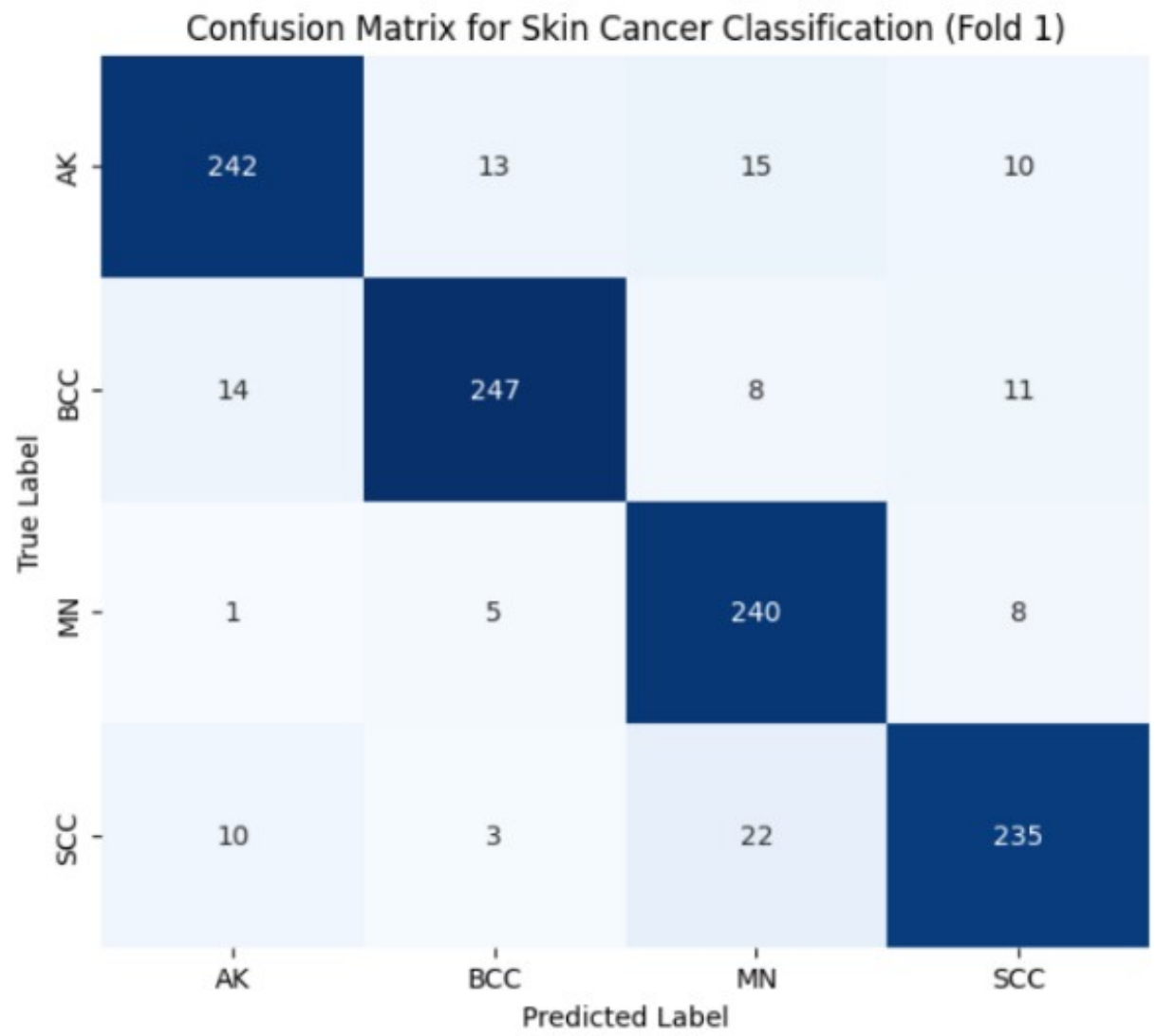

***Fig13.*** *Confusion matrix illustrating the CNN model's performance in detecting skin cancer, showcasing its accuracy and classification metrics.*

## 3.6 Colorectal Cancer:

The model was compared with existing CNN models such as VGG16, VGG19, InceptionV3, Xception, GoogLeNet, ResNet50, ResNet100, and DenseNet. The experiment concludes that VGG19 is the most effective deep-learning approach for classifying colonoscopy images. In contrast to most of its predecessors, the proposed system is fully automatic. It classifies images into the following categories: benign, benign adenomatous, moderately differentiated, and poorly differentiated malignant. The model's overall accuracy on the 165 histopathological images examined in the study was 81%.[43,44] Generally, medical images are impacted by noise or diminished quality due to blurriness; consequently, disease diagnosis is compromised. Gaussian filtering is one of the most effective filters frequently used in uniform image-processing combinations. It is crucial for noise reduction without significant distortion during this process.[45] The proposed model can merge CNN and LSTM more effectively for tumor tissue classification and cancer stage identification in medical images. This model utilizes an 18-layer CNN structure to capture the features of the pre-processed images, followed by conversion into a fully connected layer and applying the SoftMax function to achieve multi-class classification. It contains 334 images of patients with colorectal cancer, 284 of which were used to train the network, and 50 images were set aside for testing. Data augmentation aimed to increase the database five-fold. Through seven passes, this method achieved an accuracy of 94%.[46] With its 18-layer structure, the model achieved high sensitivity and accuracy, surpassing others like ANN and BPNN. Its precision, recall, and accuracy were 92%, 93%, and 91%, respectively.[47]

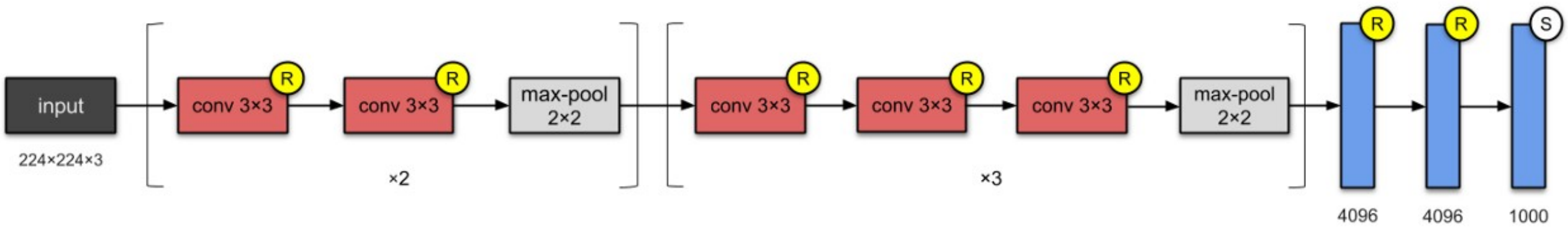

***Fig 14.*** *Architectural diagram of a CNN used for classifying colorectal cancer, highlighting key layers and connections*[48]

### 3.7 Liver Cancer:

This paper presents an ESP-UNet model for segmenting the liver in CT images that prevents under-segmentation and over-segmentation. Lightweight deep-learning CNNs have been evaluated for treating liver cancer under ALCD.[49] The LiTS dataset, containing CT liver images, was used for this study, comprising 200 CT images (130 for training and 70 for testing).[50] The proposed methodology consists of three main stages: enhancing CT liver images using a modified dual-stage Gaussian filter (MDSGF), segmenting the liver region with UNet, and detecting liver cancer using DCNN. Abdominal CT images often have low contrast and blurriness, making liver segmentation challenging. The proposed MDSGF method improves image quality by combining a dual-stage Gaussian filter with CLAHE and mean filtering, enabling more accurate liver assessment.[51]

The input image is processed through two parallel UNet networks. The first UNet performs liver segmentation, while the second UNet supplies detailed edge information regarding the liver in the CT image. Combining the outputs of these two UNets minimizes problems of under-segmentation and over-segmentation.

Various optimization algorithms, including ADAM, SGDM, and RMSProp, were utilized to train the network. The DCNN was tested with 2, 3, 4, and 5 layers and was best fitted with five layers. The precision achieved was 87.10%, 92.90%, 95.70%, and 98.60% for the 2-layer, 3-layer, 4-layer, and 5-layer DCNNs, respectively. The 5-layer model has 437,000 fewer trainable parameters than the 6-layer DCNN, which possesses 889,000 parameters. Therefore, the 5-layer DCNN was chosen. The 5-layer DCNN performs very well, achieving an accuracy of 98.60%, a precision of 0.97, a recall of 1.00, and an F1 Score of 0.98. This model was trained on the LiTS dataset in approximately 3,130 seconds. The outputs from ESP-UNet yielded the best results, with a Dice score of 0.959 and a Jaccard index of 0.921. Likewise, the DCNN model demonstrated incredibly high accuracies: at 98.60%, it showed a recall of 1.00, a precision of 0.97, and an F1 Score of 0.98.

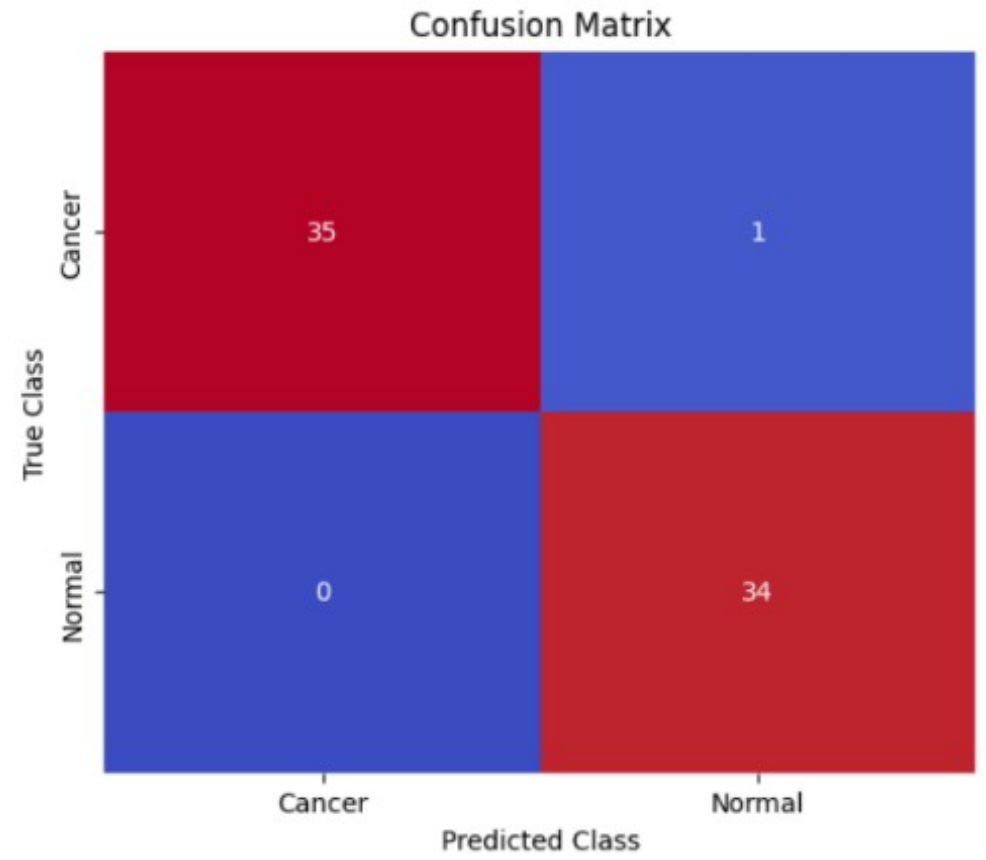

***Fig15.*** *The confusion matrix presents CNN's results in liver cancer detection, emphasizing accuracy and recall rates.*

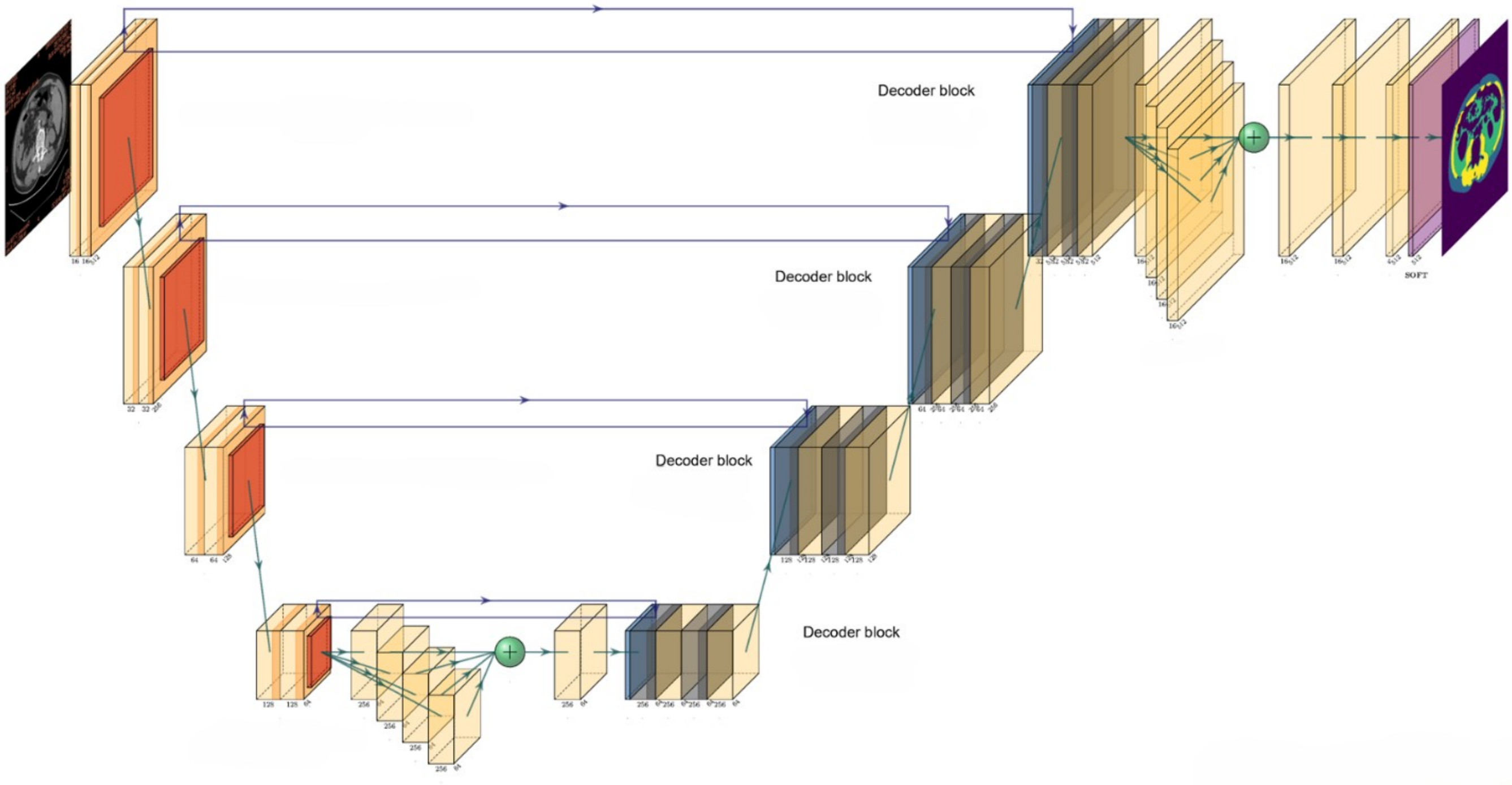

***Fig16.*** *Illustration of the ESP-UNet architecture used for liver cancer segmentation, including its compression and expansion pathways.*[52]

## 3.8 Breast Cancer:

This research utilized the natural malignant and benign images from the DDSM database, each class containing 650 images. The preprocessing involved background removal and image repair with a Wiener filter and CLAHE. This was succeeded by a wavelet packet decomposition using the Daubechies wavelet db3 at level 3, enhancing the images' smoothing.[53]

Deep convolutional networks, including AlexNet and GoogLeNet, were utilized, and optimization algorithms such as RMSprop, SGDM, and Adam were evaluated with various learning rates.[54,55]

In this method, preprocessing involved converting DDSM images to binary images, enhancing breast tissue and muscle intensity, and removing unnecessary information. The background was discarded by eliminating zero-intensity pixels, and the image was refined using Otsu's grayscale thresholding method. The noise was reduced using the Wiener filter and signal-to-noise ratio (SNR) adjustment to enhance image quality. Wavelet Packet Decomposition (WPD) was employed to eliminate non-stationary noise while preserving edges and texture by decomposing the image into sub-images across different directions.[56] The DDSM images were randomly divided into 70% for training and 30% for testing. The training was conducted using learning rates of 0.01, 0.001, and 0.0001 with SGDM, Adam, and RMSprop as optimizers.[57] Breast cancer image detection employs networks such as GoogLeNet, AlexNet, MLP, and the MLP optimized by PSO and ACO, as well as the PSO-MLP and ACO-MLP methods. GoogLeNet and AlexNet learned their accuracy and loss function values using a learning rate 0.001. Regarding GoogLeNet at epoch 1, an accuracy of approximately 98.23% was achieved with the Adam optimizer; afterward, a significant decline in fluctuations and error rate occurred. It was also observed that GoogLeNet and AlexNet exhibited very low error values near the 50th iteration for those trained using Adam as the optimizer.[58] GoogLeNet achieved the best performance with a time of 4.14 minutes and an impressive accuracy of 99%. AlexNet closely followed with an accuracy of 98.91% and a slightly longer runtime of 4.71 minutes. Next in line were the PSO-MLP and ACO-MLP, with accuracies of 90.21% and 86.14%, respectively.[59]

### 3.9 Ovarian Cancer:

Optical Coherence Tomography (OCT) imaging recordings were used to identify ovarian cancer in transgenic mice. A neural network capable of interpreting ordered spatial topographical sequences performed the classification.[60] Three neural network-based approaches were proposed: a VGG-supported feedforward network, a 3D CNN, and a Convolutional Long Short-Term Memory (ConvLSTM) network. These methods can identify significant features without requiring manual feature extraction. The comparison of the models shows that the VGG model has the highest number of learned parameters (23,121,729) but requires the least training time (201ms per sample). The ConvLSTM model has the longest training time (700ms per sample), while the 3D CNN model has the lowest number of parameters (1,140,477) and moderate training time (417ms per sample). Some challenges associated with using OCT for ovarian cancer screening include the fact that the reflected signal is nearly submerged in optical noise. In addition to the depth dependence of an imaging system's performance, the three-dimensional volumetric data reveal scale construction issues. This approach has been trained on images of mice, which differs from previous attempts. Some problems plaguing the use of OCT in ovarian cancer screening are tied to the fact that the reflected signal is almost buried in optical noise. In addition, the depth dependence of other imaging systems poses scale construction problems for volumetric data rendered in three dimensions. This approach was taught on mouse images, which differs from past methodologies.[61]

Transfer learning was applied to the VGG model, initializing it with pre-trained weights from the ImageNet dataset and fine-tuning it with the OCT data. Additionally, L2 regularization was implemented for the weights to normalize the encoder-decoder layers and dropout. In contrast, the ConvLSTM model did not utilize normalization or dropout.

The results indicated the following Area Under the Curve (AUC) values: the VGG model achieved a Peak AUC of 0.86 and a Mean AUC of 0.59. The ConvLSTM model had a Peak AUC of 0.98 and a Mean AUC of 0.81, demonstrating the best performance among the models. The 3D CNN model recorded a Peak AUC of 0.92 and a Mean AUC of 0.69. Overall, the ConvLSTM model outperformed the 3D CNN model with a Peak AUC of 0.92 and the VGG model with a Peak AUC of 0.86.

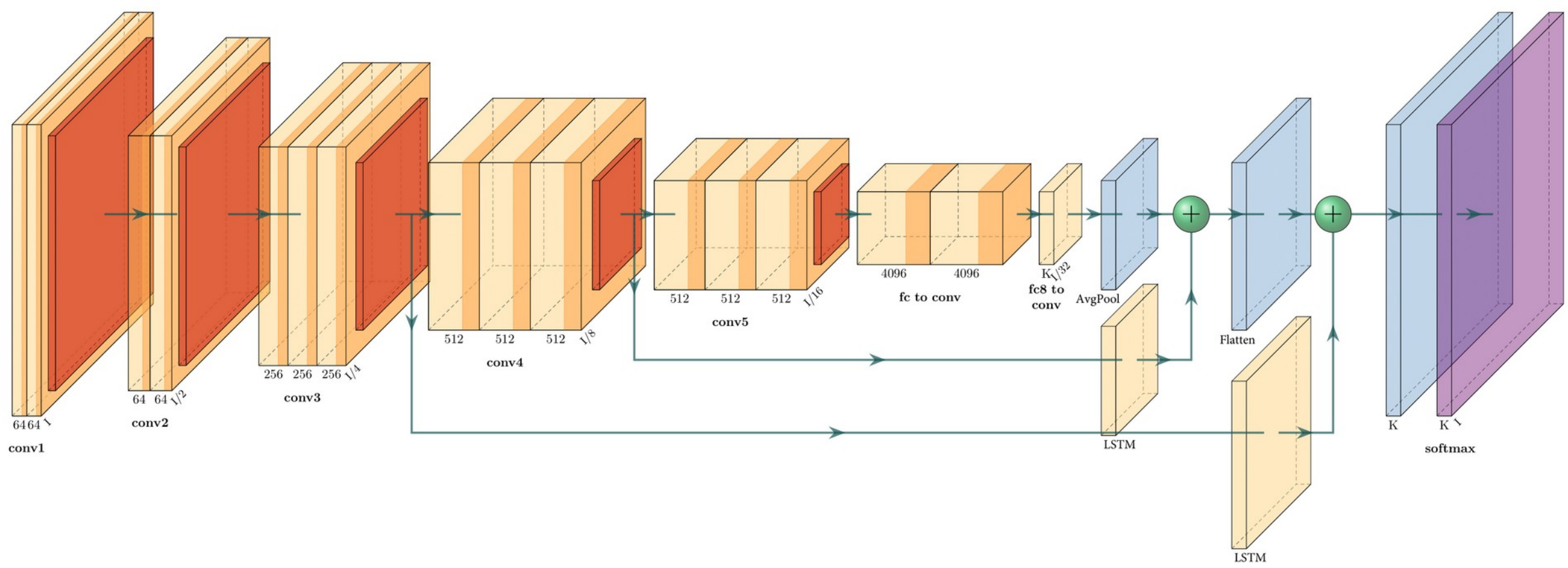

***Fig18.*** *The architecture diagram of the model used for ovarian cancer detection showcases its feature extraction capabilities.*[62]

## 3.10 Thyroid Cancer:

This study is the first to use Xception, a CNN model specialized for thyroid cancer detection.[63] The Xception model was compared with other models.

The MXTCD framework used in this study consists of two stages: In Stage 1, medical images are fed into the Xception model for binary classification. This stage evaluates the impact of the type of medical images (CT and ultrasound) on diagnostic results, with model tuning applied to different datasets. In Stage 2, multi-channel architectures based on Xception are used for binary and multi-class classification tasks tailored to the needs of clinicians. Three optional architectures were proposed: the Single Input Dual Channel (SIDC) combines input channels into a unified model; the Dual Input Dual Channel (DIDC) combines dual inputs for more refined classification; and the Four-Channel Architecture processes four input channels simultaneously for high accuracy. Xception is a simplified description of Inception, using depth-wise separable convolution to recognize channel and spatial correlations. It outperformed other contestants in the ImageNet competition, the models VGG16 and ResNet152, in accuracy performance and computation time.[64]

The supplementary dataset included 448 DDTI ultrasound images, divided into 66 benign and 382 malignant images; 917 Hospital_X ultrasound images, comprising 200 benign and 717 malignant images; and 2,352 CT scan images from Hospital_X. After excluding 577 non-teaching images, this left 578 benign and 514 malignant CT images. The ultrasound images were labeled according to TIRADS scores, while the CT images were labeled based on histopathological results. A thyroid segmentation tool in Python was utilized for cropping and labeling the images. All images were rescaled to a standard size of 224 × 224 pixels.[65]

Xception performed better than all other models, achieving an accuracy of 0.980 for Ultrasound DDTI, 0.987 for Hospital_X Ultrasound, 0.966 for CT scans, and 0.970 for CT scans. Xception also stands out with extremely high values for predictive accuracy, NPV, recall, and F1 score. Although ResNet10 was the fastest run, the running times for Xception and DenseNet121 were close. In this work, multi-channel architectures were researched, and the optimization was primarily related to the sizes of filters for SIDC. Specifically, 3 × 3 and 7 × 7 filter sizes were examined, where the latter gained better accuracy, with accuracies of 0.984 for DDTI ultrasound, 0.988 for Hospital_X ultrasound, 0.972 for left CT scans, and 0.974 for proper CT scans. SIDC demonstrated a slight improvement in other models for all datasets; for example, the accuracy for DDTI U/s increased with SIDC from 0.984 to 0.987. With an average accuracy of 0.95, DDRI fares better against the four-channel model, which has an accuracy of 0.94. The four-channel model shows a slight advantage for "normal" patients (accuracy = 1.00), while the DIDC performs better concerning "abnormal" ones.[66]

Unlike previous ultrasound studies that reported diagnostic accuracy of 70% to 92%, this study found that DDTI ultrasound achieved an accuracy of 0.980 compared to 0.987 for the Hospital_X ultrasound. Earlier CT studies indicated accuracy levels ranging from 90.4% to 95.73%. Results suggest that the performance of the Xception model surpasses that of its multi-channel architectures when compared to single-channel CNNs alone. Thus, this finding exceeds the state of the art in thyroid cancer detection, particularly with the inclusion of CT scans.

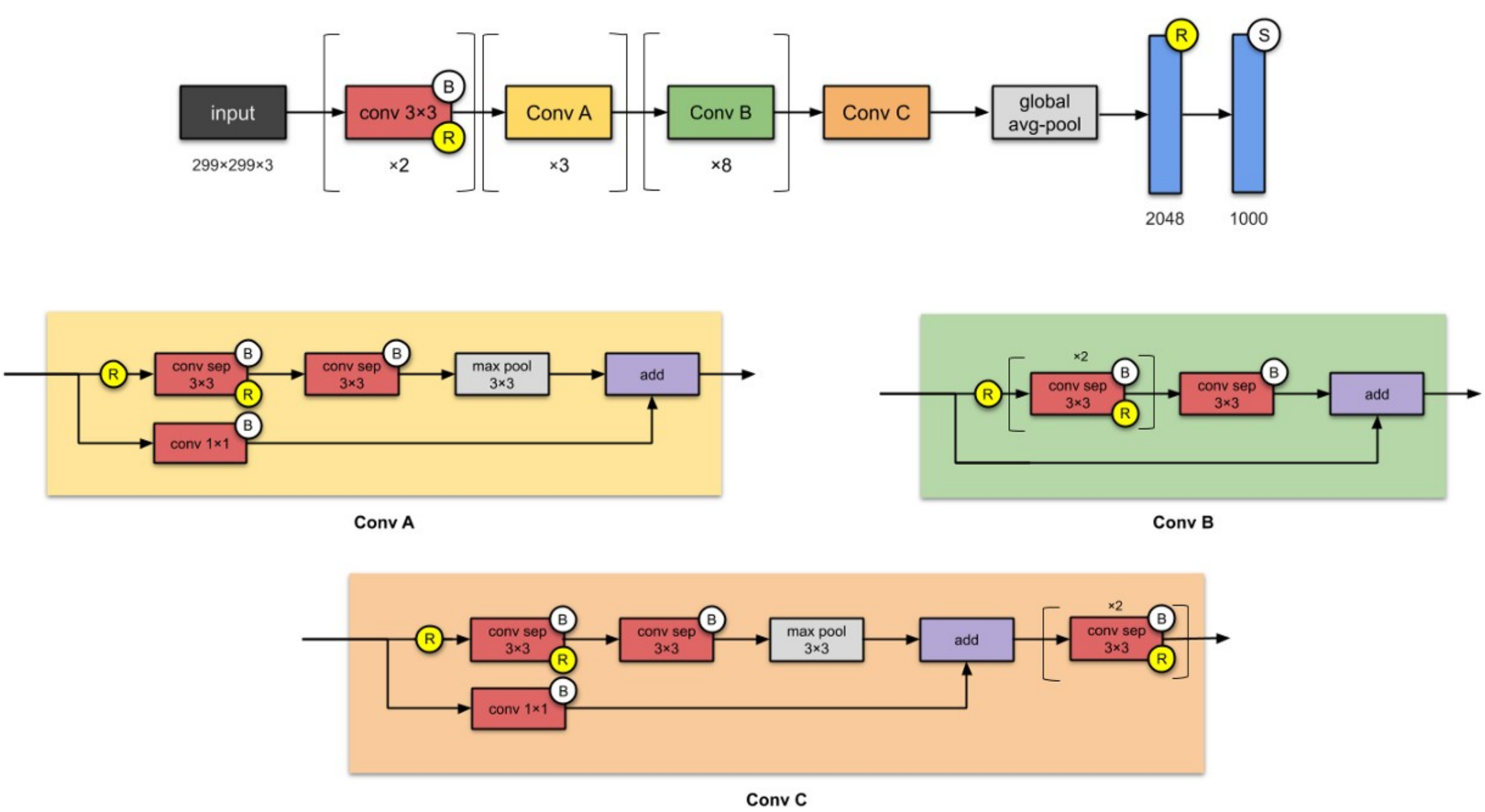

***Fig19.*** *Architectural representation of the Xception model optimized for thyroid cancer detection, highlighting its depth-wise separable convolutions and multi-channel architecture.*[67]

In Table 2, we have extracted and comprehensively presented the key details and information of each of these 10 articles. The important information in these articles includes aspects such as accuracy, the type of dataset used, and most importantly, the neural network architecture employed. Additionally, the table allows us to identify the strengths and weaknesses of each method of detection, analyze parameters, examine specific techniques applied to images, and overall, use it as a quick and practical reference.

*Table 2. Comprehensive comparison of the articles reviewed above, highlighting the key aspects of each CNN mode*

| Reference | Cancer Type | Dataset Used | Imaging Technique | CNN Architecture | Accuracy (%) | Loss Function | Optimizer |
|---|---|---|---|---|---|---|---|
| 68 | Breast | DDSM | Mammography, Ultrasound | AlexNet, GoogLeNet | 99.0 | Binary Cross entropy | SGDM, RMSprop, Adam |
| 69 | Lung | IQ-OTH/ NCDD | CT scans | AlexNet | 93.54 | Binary Cross entropy | Adam |
| 70 | Colorectal | TCGA-CRC-DX | Histopathology | VGG19 | 92.0 | Cross entropy | Adam, SVM classifier |
| 71 | Thyroid | DDTI, CT Hospital_X | Ultrasound | Xception | 97.5 | Dice Loss | Adam |
| 72 | Ovarian | OVCAD | OCT | VGG16-LSTM Hybrid | 98.0 | Mean Squared Error (MSE) | SGD |
| 49 | Liver | LiTS | CT scans | U-Net | 98.6 | Binary Cross entropy | Adam |
| 73 | Bladder | Custom Dataset | Cystoscopy | GoogLeNet | 98.0 | Cross entropy | Adam |
| 74 | Leukemia | ALL-IDB, C-NMC | Microscopy Images | BCNN-LSTM Hybrid | 94.0 | Binary Cross entropy | RMSprop |
| 75 | Prostate | PROMISE12 | MRI | GoogLeNet, SVM | 100 | Binary Cross entropy | Adam |
| 76 | Melanoma | ISIC | Dermoscopy Images | Custom, MobileNet, GoogleNet | 95.98 | Binary Cross entropy | Adam |

## IV. Conclusion:

CNNs are among the most developed methodologies identified from review results for assessing medical images to detect a wide range of cancers. The performances recorded in the diagnostic models from the reviewed literature were outstanding, mostly over 95% and up to 99% in some cases. For instance, GoogLeNet ran in 4.14 minutes with 99% accuracy, whereas AlexNet took 4.71 minutes but still had equally high accuracy at 98.91%.

For thyroid cancer, the range for Xception lies between 96.6% and 98.7%, depending on whether ultrasound or CT scans are in question. ConvLSTM achieved the best results in ovarian cancer with accuracy and AUC of 98% and 0.98, respectively, whereas other deep models like VGG have results touching AUC as low as 0.86. In the case of skin cancer, fine-tuning using augmented data resulted in a model accuracy of 95.98%, whereas that by GoogLeNet and MobileNet was 94.22% and 94.86%, respectively.

This is further enhanced by transfer learning using GoogLeNet, which classifies data with 99.71% accuracy while the data is limited; the dataset contained 600 samples only for prostate cancer. Then, given CT scans of liver cancer, ESP-UNet made predictions of cancerous areas with 98.6% accuracy. Some of the preprocessing techniques adopted in this work include Gaussian filters, CLAHE techniques, and wavelet decomposition, all returning low noise and improvement in quality.

Other limitations have been imposed due to the unavailability of human data in some works, mainly those related to ovarian cancer, where mice yielded data. Several techniques used in the process include data augmentation and transfer learning, which are under effective control in such situations. Techniques such as dropout and L2 regularization, while avoiding overfitting, helped to enhance accuracy with reduced errors.

It finally points out a few conclusions: With more good quality available data, further improvements are always possible with enhanced pre-processing methods and applying different types of sophisticated CNN architectures to have better accuracy and speed of diagnosis. With such excellent prospects, this technology of CNN can be transformed into one of the enabler tools in the clinical system at large to make cancer diagnosis less time-consuming and less expensive in the years to come.